# Robotic capabilities framework

A boundary object and intermediate-level knowledge artifact for co-designing robotic processes

**Authors**
Alessandro Ianniello* [1,2,4]
Dave Murray-Rust [2]
Sara Muscolo [3]
Olger Siebinga [2,4]
Nicky Mol [4]
Denis Zatyagov [5]
Eva Verhoef [5]
Deborah Forster [5]
David Abbink [2,4]

[1] Department of Architecture and Design, Politecnico di Torino, Turin, Italy
[2] Industrial Design Engineering, TU Delft, Delft, The Netherlands
[3] Independent researcher, Italy
[4] Mechanical Engineering, TU Delft, Delft, The Netherlands
[5] RoboHouse, Delft, The Netherlands

*Corresponding author: alessandro.ianniello@polito.it

# Abstract

As robots become more adaptable, responsive, and capable of interacting with humans, the design of effective human-robot collaboration becomes critical. Yet, this design process is typically led by monodisciplinary approaches, often overlooking interdisciplinary knowledge and the experiential knowledge of workers who will ultimately share tasks with these systems. To address this gap, we introduce the robotic capabilities framework, a vocabulary that enables transdisciplinary collaborations to meaningfully shape the future of work when robotic systems are integrated into the workplace. Rather than focusing on the internal workings of robots, the framework centers discussion on high-level capabilities, supporting dialogue around which elements of a task should remain human-led and which can be delegated to robots. We developed the framework through reflexive and iterative processes, and applied it in two distinct settings: by engaging roboticists in describing existing commercial robots using its vocabulary, and through a design activity with students working on robotics-related projects. The framework emerges as an intermediate-level knowledge artifact and a boundary object that bridges technical and experiential domains, guiding designers, empowering workers, and contributing to more just and collaborative futures of work.

# 1. Introduction

Robots are increasingly being introduced into workplaces to take on tasks that are dull, dirty, and dangerous for humans [3]. While this shift can relieve workers from such risks and hindrances, it may also lead to negative experiences, such as boredom or a sense of dehumanization in daily routines [27].

Furthermore, the integration of robots into the workplace is driven by the need to address complex challenges, such as labor shortages and the request for increased productivity [60]. At the same time, implementing robots into complex work environments raises broader socio-technical and ethical challenges, including questions of social justice, shifts in power dynamics, changes in work practices, and the adaptation and acceptance of technology by workers [64]. In order to address the complexity of these issues and strive for more just and meaningful forms of work [56, 75], there is a need to investigate current and potential relations between workers and robots [75]. Thus, we see the value in developing frameworks that can support the design of robots, and their capabilities, which, in turn, should be able to complement human skills, and realize meaningful, collaborative arrangements between humans and robotic platforms [31].

Realizing this vision requires rethinking approaches to designing human-robot collaboration, which in turn requires a collaborative engagement with what robots are imagined to be. It calls for approaches that span multiple disciplines and non-academic knowledge domains, bringing together technological insight, worker experience, and design principles. In other words, shaping meaningful human-robot collaborations might benefit from a transdisciplinary approach. Transdisciplinarity (see among others [23, 44, 50, 51]) is a process of knowledge creation through the collaboration of multiple disciplines, non-academic stakeholders, and social groups, that fosters new learning practices to address complex socio-technical issues [23, 36] and generate positive social change [74].

A continuing challenge with transdisciplinary working is supporting clear and active communication between various stakeholders [47, 69]: between different disciplinary knowledge domains which might not speak the same language; and between academic and non-academic stakeholders who might have different levels of understanding of a specific technology or issue.

In our case, beyond general questions of transdisciplinary science communication, we are interested in how to create knowledge that crosses the boundaries [1, 35] between different stakeholder groups and academic communities. As we need to involve workers who have little knowledge of robotics in both coproducing solution-oriented and transferable knowledge [56] and working from conceptual frameworks towards applying research findings to address real-world problems [36], there is a need for representations that support communication between roboticists and non-experts in the co-design of interventions. The aim of this work is to develop a framework in this space to create work that is not only technically informed but also socially and ethically relevant, becoming a tool to achieve more just and meaningful worker-robot relationships [56, 75]. Given these considerations, we pose the following research question:

**RQ:** "*How can we create a space where robotic and worker capabilities can be compared, mixed and played with in support of meaningful collaboration in the context of the future of work?*"

To answer this question, we co-developed a *robotic capabilities framework* that captures aspects of robotic technology in a comprehensible manner while maintaining technical rigour. We view this framework as both a boundary object [61] – a shared reference that different stakeholders can use to communicate and collaborate – and as a form of intermediate-level knowledge [41] that bridges concrete design insights with generalized understanding. By making robotic capabilities transparent and accessible, the framework can inform workers, designers, engineers, and researchers about what current robots can and cannot do. The goal of this framework is to foster collaboration rather than substitution, by creating a space where worker's and robot's capabilities can be negotiated and mixed. In turn, this should allow us to envision work arrangements where humans and robots work together, each contributing what they do best, toward more meaningful futures of work.

The contribution of this paper is a framework for describing robotic capabilities, which comprises the following:
1. A grounding of *robotic capabilities* as actions and activities that robotic systems may be able to carry out relative to a task and a socio-physical context.
2. The development of a taxonomy of these capabilities based on literature research and reflexive discussion among both roboticists and designers, creating terms that are both technically rigorous and understandable, thereby serving as a common reference point for transdisciplinary conversations.
3. The application of the taxonomy, to:
   a) Describe robots, demonstrating both simplicity and the ability to differentiate important qualities.
   b) Demonstrate its ease of use and ability to quickly support meaningful discussion by non-experts, in our case with groups of interdisciplinary students.

The remainder of the paper is organized as follows. First, we provide background on related work. Next, we detail how we created the robotic capabilities framework and we present it. We explore and evaluate the framework against a set of criteria to ensure that it is fit for purpose, involving experts and design students in two different activities. In the discussion, we link our framework to the concept of boundary objects and intermediate-level knowledge, and reflect on the role of diverse knowledge in shaping our work, as well as its limitations. Finally, we conclude by summarizing key insights, highlighting how our contributions advance the current debate on the future of work, and suggesting directions for future research.

# 2. Background

## 2.1 Research context

The research presented here is part of a broader research, in which several directions are pursued, with the goal of gradually applying a transdisciplinary approach (see Section 2.2.1) to its overall development. The overarching aim of this broader research is to gain knowledge and insights around how to effectively enhance workers' well-being and improve the overall quality of work when implementing robots in existing work processes.

The main stakeholder of the project is an organization that operates in multiple aviation sectors. The organization is researching and testing how to integrate robots and other technologies across various operational departments, such as MRO, to address broader challenges – for instance, labor shortages and the demand for increased productivity – and specific risks that may compromise workers' physical and psychological wellbeing. In this context, in fact, expert workers frequently perform heavy physical tasks that span from metal working to (dis)assembly operations.

As part of the project's transdisciplinary endeavour, and in recognition of the socio-technical complexity of integrating robots into sensitive work contexts, a design researcher (the first author) was included in the team to explore how design could further contribute to shaping responsible and worker-centered innovation.

During the development of a design-driven activity, several challenges emerged within our research group regarding the relevant concepts and ideas to be leveraged in this process. These tensions initiated discussions about the necessity of establishing a common language regarding robots and their capabilities, among the diverse fields represented in our group, namely robotics engineering, with a background in human factors and man-machine systems; design researchers, with a background in speculative and interaction design; and practitioners, with an expertise in mechanical, robotics, and software engineering, and design. This is necessary to find an internal consensus for discussing collaborations between workers and robots. This common language also aims to effectively and meaningfully engage workers in design processes by equipping them with the relevant technological knowledge.

## 2.2 Related works

### 2.2.1 Transdisciplinary approaches to Human-Robot Interaction

The field of HRI is experiencing a move from disciplinary silos toward the creation of more inclusive modalities of knowledge production [73]. This move is further pushed by the realization that designing robotic solutions for the real-world requires multistakeholder approaches that integrate different forms of knowledge [19]. When robots stop to be considered only as technical systems, to be investigated through laboratory studies, they consequently become part of socio-technical systems [8]. This leap brings with it several onto-epistemic challenges around what robots are, should be, and how they should relate with humans [57].

In this context, transdisciplinarity [50, 51] emerges as a promising approach to knowledge production that can unpack this newly discovered complexity around HRI. Transdisciplinarity involves interdisciplinary and multistakeholder collaborations, recognizing that complex real-world issues can't be addressed by a single discipline or community [19, 48, 50, 68]. It should further promote awareness of the different stakeholders, and reflexive competencies of the people affected by and involved in the processes [5] enabling them to define research questions, briefs, and approaches.

In the context of HRI for the future of work, this means combining technical robotics expertise with insights from design, social sciences, work psychology, ergonomics, organizational studies, and the tacit knowledge of workers on the ground. Prior works have argued that the HRI field must work beyond disciplines and include plural perspectives through transdisciplinary research done with and for workers [56, 75]. This inclusive mode of research is essential for envisioning and realizing new forms of human-robot collaborations that are socially acceptable and equitable [74]. By engaging diverse perspectives, transdisciplinarity can identify blind spots that a single-discipline approach might overlook [24].

While transdisciplinary collaboration offers great potential, it also faces challenges in reconciling diverse epistemic cultures and developing a common language [19, 47, 69, 70]. Transdisciplinary approaches must be understandable and reliable for experts and non-experts, and instigate conversations between practitioners, academics, and people alike [45]. That's why, when doing transdisciplinarity, we need intermediate-level knowledge [41] and boundary objects [61] to allow interdisciplinary and multistakeholder collaborations.

The development of the robotic capabilities framework follows this approach: it was informed by collaborations between robotics engineers, design researchers, and practitioners. This blending of expertise ensured that the framework's content and form would be relevant and understandable across different communities.

### 2.2.2 Intermediate-level knowledge and boundary objects

The HRI field is increasingly drawing on intermediate-level design knowledge [20, 25, 41, 42] and boundary objects [1, 7, 38, 43, 61] to bridge the gap between abstract theory and situated practice.

Intermediate-level knowledge is a middle territory of knowledge, which is more generalized than a single design instance, but more situational than theory [41]. Examples of intermediate-level knowledge include, for instance, design methods and tools, design guidelines, patterns, heuristics, documented experiential qualities, critical analyses, strong concepts, and annotated portfolios [20, 25, 41]. These forms of knowledge share an ability to capture transferable insights that practitioners and researchers can apply in new projects and research. Such intermediate-level contributions are valuable because they allow the community to accumulate knowledge in a form that directly informs design and research. Because of its tangible, practical nature [25], intermediate-level design knowledge can offer a valuable contribution to a transdisciplinary endeavour.

On the other hand, the concept of boundary objects is derived from sociology and design. Originally introduced by Star and Griesmer, it describes artifacts that are "plastic enough to adapt to local needs… yet robust enough to maintain a common identity across sites" [61, p.393]. Boundary objects can act as levers to mediate collaboration across different communities and disciplines. They can be seen as common reference points that every stakeholder can interpret through their own lens, facilitating communication and alignment [49]. They are not fixed yet adaptable to local needs, while maintaining a certain degree of familiarity across the involved communities and disciplines [49]. In HRI, examples of boundary objects include prototypes, sketches, scenarios, storyboards, and other design artifacts that stakeholders can discuss, rely on, and manipulate, according to both the situated and general knowledge at play in the processes. Like intermediate-level forms of knowledge, boundary objects serve well transdisciplinary endeavours, ensuring that roboticists, designers, workers, and other stakeholders can co-create the robot's role. By relying on boundary objects to mediate transdisciplinary collaborations, organizations can better account for the tacit, situated knowledge of workers, resulting in robots that fit more naturally into their environments.

### 2.2.3 Representing robots in Human-Robot Interaction

Robotic capabilities were primarily discussed through hardware-centric descriptions, emphasizing technical aspects such as mobility and manipulation. Early studies, for instance, formalized these capabilities, establishing foundational technical models for robotics [33, 53].

Over time, a shift towards comprehensive, multidimensional frameworks integrating physical, cognitive, and social dimensions took place [39, 40]. Such frameworks increasingly recognize robots as multifaceted entities capable of autonomous task execution and cooperative skills, reflecting their growing complexity and integration into human environments.

Recent literature often employs taxonomic approaches, delineating robotic capabilities into specific dimensions such as mobility (navigation, motion primitives, localization), manipulation (pick-and-place, assembly tasks), and interaction (human-robot communication, social cues, expressive behaviors). For instance, the frameworks proposed by Linjawi and Moore [40] and Hu et al. [28] categorize robots based on these clearly defined dimensions, addressing both technical precision and social complexity.

Within HRI specifically, the conceptualization of robots varies widely. Robots are frequently described as technical tools reflecting their functional role [55]. Conversely, other narratives position robots as social companions, actors endowed with anthropomorphic and emotional qualities, emphasizing relational interactions and social companionship [34, 46, 71].

In other words, technical approaches to HRI typically focus on what a robot *can* do, given its technical characteristics, whereas social approaches to HRI concentrate on what a robot *should* do within a human environment. Path planning provides a clear example of this distinction. In the work of de Grot et al. [15], the authors present a method to optimize path planning while constraining the probability of collisions with obstacles during motion. In contrast, Saerbeck and Bartneck [58] examine how robot motion influences the perceived affective state of the robot, deriving "design knowledge for the design of movement behaviors of social robotic interfaces" [58, p.60]. Finally, Francis et al. [18] propose a framework to understand not only how a mobile robot can and should operate around humans but also how humans respond to its presence, behaviors, and actions. This framework represents an effort to bridge technical and social perspectives, offering a more comprehensive view of robots and their capabilities within HRI.

Overall, the evolution from hardware-centric descriptions to nuanced, multidimensional frameworks underscores an ongoing redefinition of robotic capabilities. This progression aligns with broader societal shifts, advocating for more equitable and inclusive human-robot collaborations within workplace contexts.

Despite these significant advancements, some gaps remain. In particular, systematic ways to operationalize and compare robotic capabilities across different settings are still emerging, and the challenge of reconciling divergent disciplinary perspectives continues to hinder the development of shared, durable languages for collaboration.

## 2.3 Synthesis and research gap

The reviewed literature points to three main developments in HRI. First, transdisciplinarity has been identified as a promising way to address the socio-technical complexity of HRI. By combining technical expertise with perspectives from design, social sciences, ergonomics, and workers' tacit knowledge, it enables more inclusive and equitable forms of collaboration. However, operational challenges persist, particularly in reconciling epistemic differences and creating shared languages that make such collaboration sustainable and meaningful.

Second, intermediate-level knowledge and boundary objects have been proposed as valuable tools to connect abstract theories with situated practices. While they show potential to mediate transdisciplinary collaborations, their systematic role in HRI is still underexplored.

Third, representations of robots have shifted from hardware-centric accounts toward multidimensional frameworks that integrate physical, cognitive, and social dimensions. Yet, practical methods, tools, and frameworks for assessing and comparing these capabilities across contexts remain underdeveloped.

Together, these developments reveal a central research gap: although the HRI field increasingly calls for multidimensional and participatory approaches, it lacks concrete, operational frameworks that can act as boundary objects to support collaboration across diverse stakeholders. This paper addresses this gap by introducing the worker–robot capabilities framework as an intermediate-level contribution designed to mediate transdisciplinary practices in the context of robots at work.

# 3. Framework Construction

To introduce our framework, we present i) the overall methodology we applied to define it, and the design criteria derived from the research context, which the framework should meet (Section 3.1); ii) our working definition of a robot necessary to establish an initial common ground among our diverse disciplinary knowledge (Section 3.2); iii) an encompassing 'Task' object (Section 3.3) that allows connection to existing human activities; and finally iv) the groups and categories of the framework itself (Section 3.4). Together, this gives a framework and an associated HRI vocabulary that allows the integration of robotic capabilities in collaborative work environments.

## 3.1 Methodology and design criteria

Our framework has been constructed and refined by a group of academics and industry practitioners that includes expertise in robotics, engineering, HRI, design, participatory and transdisciplinary practices. Over an initial period of approximately four months, the first author conducted two focus groups with the other members of the research group and multiple one-on-one discussions. In both these types of sessions, we discussed our diverse definitions of robots and what they should be capable of to be considered as such. These sessions were recorded, transcribed, and supplemented with additional insights contributed through a shared document.

We then analyzed all the materials we collected, identifying points of convergence and divergence, which we synthesized into: i) a working definition of robots; ii) and the first draft of the framework.

By conducting a comparative analysis with existing literature, we evaluated the framework's relevance and soundness. We then refined it through constant linguistic iterations until we reached its final draft, which we evaluated against design criteria developed beforehand.

To test the validity of the framework, we asked experts to use it to describe existing robotic platforms (Section 4.1), and we carried out a workshop with students (Section 4.2) during which we introduced them to an adapted version of it. In Figure 1 we show the overall process we followed to develop our framework.

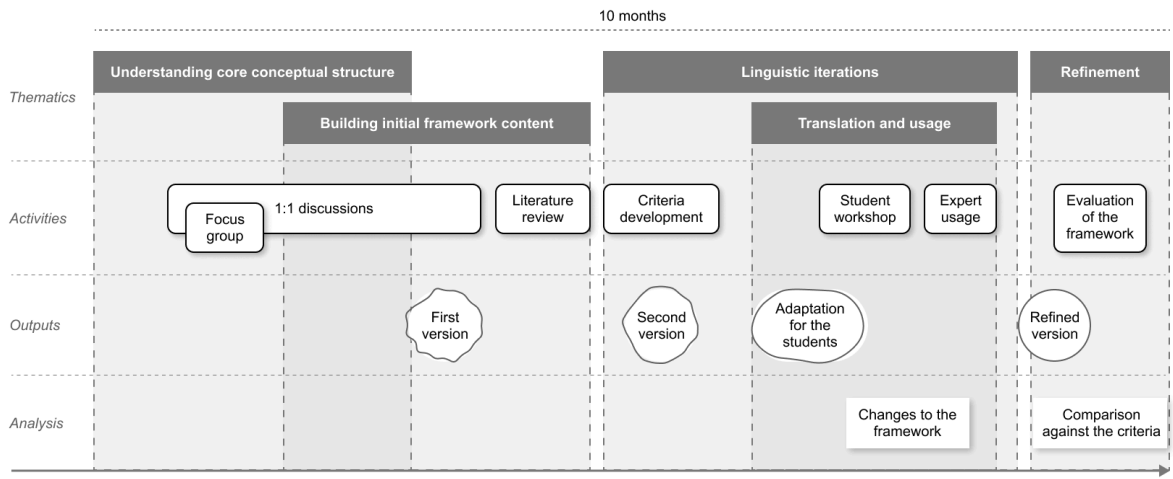

Fig.1 Flow of the process followed to conceptualize, develop, and validate the framework. In the top part of the picture we report the thematics addressed throughout the process. Moving from top to bottom, we then find the specific activities we carried out, the outputs of each iteration, and the analytical level.

Below, we explain our design criteria, making explicit their rationale:
1. *The framework captures the range of functions that current and near-future robots may offer, while presenting them in a simplified, accessible format that is understandable even to non-experts, and avoiding technical specification* (**C1**).
   This criterion focuses on comprehensiveness and accessibility, ensuring that the framework is multidimensional and allows comparison of the capabilities across different contexts, while being understandable for different stakeholders.
2. *It serves as a boundary object, supporting high-level, transdisciplinary discussions about the integration of robots in work contexts, and facilitating communication and collaboration between different stakeholders* (**C2**).
   This criterion emphasizes mediation, positioning the framework as a tool that connects different disciplines and communities, helping to reconcile epistemic differences and foster shared languages.
3. *It emphasizes what robots can do, highlighting distinctive features that help compare and differentiate robots in meaningful ways* (**C3**).
   This criterion highlights the need to view robots through their capabilities and impact on work practices to build common ground across stakeholders and compare robots that may otherwise seem alike.

In Section 5.1, we return to these criteria to evaluate the framework and examine whether any aspects were overlooked during its development and testing.

## 3.2 Working definition of robots

While the literature presents a rich body of knowledge regarding what a robot is (see among others [11, 22, 46, 54, 59, 63, 76, 77]), there appears to be a lack of consensus on a clear definition. Since we want to create a framework that allows people to think broadly about what robotic technologies should do, it is very useful to have a sufficiently clear and open definition of robot to work with.

Given this situation, and considering that our research group comprises individuals with diverse academic and nonacademic perspectives, we needed to align on a common definition of robots. By doing so, we enabled us to further define the capabilities of robots.

Through the process of mediating and negotiating among our diverse perspectives, we define robots as "*physical, mechanical, and artificial agents that can sense the world, the environment, other agents, and objects; reason to make actionable decisions through the development of plans; change their behavior based on their sensing and reasoning (and vice versa); and interact with the physical world, the environment, other agents, and objects to purposefully perform actions*". We chose to define robots as "agents" to emphasize the relational and complex aspects of human-robot interaction. At the same time, to prevent misunderstandings about the nature of robots, we maintained a focus on their "physical, mechanical, and artificial" characteristics. Lastly, as a point highlighted by all the knowledge areas, we clearly mentioned the action-based and performative goals of robots.

## 3.3 Framing capabilities in the context of tasks

Our framework is intended to allow the design of scenarios in which human activity and robotic technology are intermixed. In order to support this, we frame our framework around tasks, where a particular outcome is desired, but it may be achieved by multiple means.

We chose to focus on the task dimension, recognizing several important reasons for this decision. Work is typically structured around tasks of varying complexity and interdependence [4, 26]. Tasks are the fundamental units of coordination, providing the lens through which collaboration and articulation of work are enacted [62]. In HRI, task allocation has similarly been recognized as a central challenge, where the key issue is deciding who does what, when, and how [52]. It is within this dimension that both workers and robots exercise the capabilities necessary to perform activities and actions, and this is the primary context where most relationships between them are established [13].

This focus serves as a crucial lever for engaging workers, as it is through the task lens that their knowledge, expertise, and experience can emerge. Participatory and ergonomic approaches to work design have long emphasized the centrality of tasks as the ground where human capabilities, constraints, and know-how are expressed [32]. Moreover, research has shown that mismatches in task allocation between humans and machines can produce unintended and unwanted consequences [6]. By centering our work on the task dimension, we foreground not only efficiency but also workers' lived experiences.

Lastly, focusing on tasks creates a space where workers can actively envision and imagine possible and preferable relationships with robots, because tasks are directly tied to their everyday practices and experiences. This resonates with work on mixed-initiative and adaptive task allocation, which highlights that division of labor between humans and machines is not static but negotiated, context-dependent, and shaped by trust, expertise, and situational demands [66, 67]. Such a framing can enhance workers' agency and empower them to express their desires, expectations, needs, and concerns, giving them a voice in shaping robotization processes in the workplace.

## 3.4 Capabilities

The definition of robots provided above helped us establish a common ground for discussing their capabilities [10, 29, 39, 72]. Our emphasis on capabilities arises from the desire to analyze the work context in terms of the functional units necessary for accomplishing a task. By adopting this perspective, we can focus on the specific capabilities of robots required to complete a task, without the need for an exhaustive breakdown of all of them.

Furthermore, by exploring the capabilities of robots within the task dimension, workers can more readily relate these to their own capabilities, envisioning what a constructive relationship might look like and identifying opportunities for enhancing their work practices, further overcoming potential concerns.

We broadly define capabilities as "*potential abilities that an agent possesses and can exercise, with their realization influenced by specific conditions*". This definition applies to both humans and robots, aligning with our objective of establishing a framework where the capabilities of workers and robots can be compared, exchanged, and negotiated within the task dimension and context.

Within our group, the robotics engineering perspective emphasized that a robot should possess the abilities to "*sense the world*", "*reason*", and "*change its behavior based on its sensing and reasoning*". Furthermore, robotics engineers stated that robots should be capable of "*interacting with the environment*". This set of capabilities resonates with the characteristics we mentioned in the definition of robots. They further emphasized the importance of concentrating on "*actions that humans can or cannot perform to achieve an expected result*".

Like the robotics engineering perspective, the design perspective emphasized the importance of defining capabilities within a "*task space*" and "*in terms of the target activity*". This approach allows the "*purpose*" of the capabilities to emerge, highlighting the "*behavioral and social aspects*" that underpin potential relationships between workers and robots. The design perspective further offered an insightful way to semantically frame capabilities, suggesting they should "*combine a verb and an object*" to create a grounded class of possible actions.

However, during this process, some friction arose around the use of the term "*capabilities*" which was proposed by the design researchers. The robotics engineering researchers favored instead terms like "*technological possibilities*". The term technological was employed to denote that a given feature may also be found in other technologies, whereas the term possibility was used to underscore that a particular feature can be present in, or implemented within, a robot but not necessarily in all robots. Moreover, some of the engineers' concerns centred on the need to delineate the boundaries of how a given problem can be roboticized, since robots can be designed with a wide range of capabilities.

Conversely, such choices would have raised the level of abstraction to the point of undermining both the meaning and the specificity required for the aims of this study.

These tensions were addressed by clarifying the rationale for selecting the term "*capabilities*." As noted earlier, this perspective places emphasis on tasks, which in turn may help workers and other non-experts feel engaged and act as active contributors in imagining technological implementation. Put differently, because our aim was to establish a shared, transdisciplinary vocabulary for envisioning work environments with robotic platforms, the term "*robotic capabilities*" proved more consistent with this objective.

What we present is the final version of the framework (Fig. 2) that was iterated through different activities (see Section 3.1).

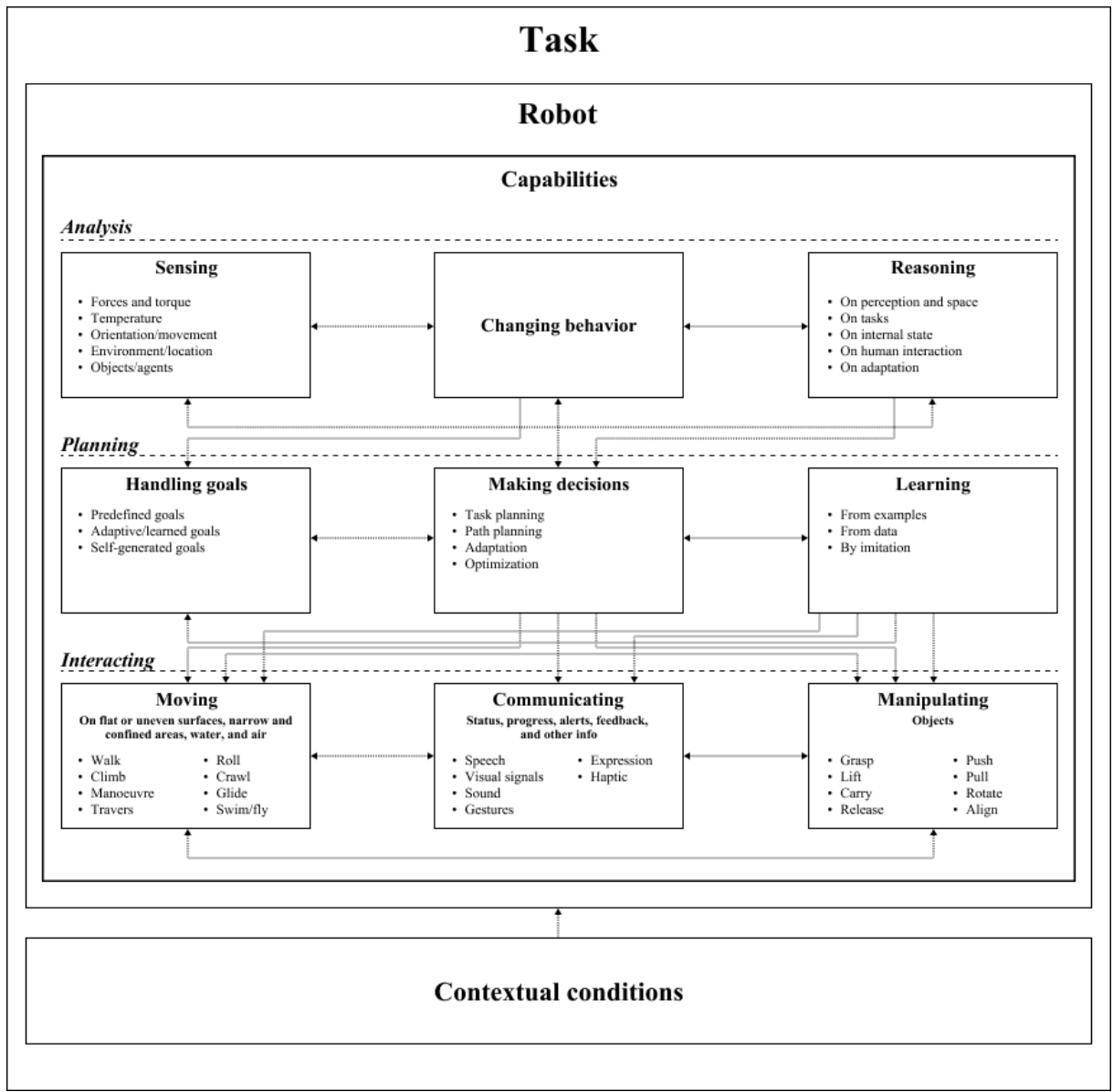

Fig. 2 The robotic capabilities framework, that identifies general robotic capabilities within a task dimension and depending on contextual conditions. It is divided into three different layers: i) analysis, ii) planning, and iii) interacting.

We categorized robotic capabilities into three distinct layers:
- Analysis: here, we find the capabilities that enable robots to sense and analyze their surroundings and other agents within the space [39, 72]. The exercise of these capabilities is neither physical nor visible. In this first layer, we identified sensing capability, i.e. the ability to perceive the surroundings, objects, other agents, movements, temperature, forces, and torque through different types of sensors [10, 29, 59]; reasoning capability, i.e. the ability to process information on perception and space, tasks, internal state, human interaction, and adaptation [2, 13]; and changing behavior capability, i.e. the ability to modify behaviors based on sensing and reasoning inputs, and vice versa [14, 37, 72].
- Planning: this layer encompasses the capabilities that allow robots to formulate a plan of actions to complete a task [2, 10]. In this case as well, the exercise of these capabilities remains largely non-physical and invisible, with the exception of the learning capability, which can manifest in the physical dimension. Robots can handle pre-defined, adaptive and learned, or self-generated goals [29, 37]. They can make decisions to achieve those goals, especially in terms of task, motion, adaption, and optimization [10, 14, 30, 37]. They can further learn from example, data, or by imitation, enabling them to refine and improve the action plan over time [2].
- Interacting: the final layer consists of the robot's interactive capabilities, which are the ones that most directly influence the relationship between workers and robots. We employed a general framework that categorizes these interactive capabilities into three main areas. Moving within different types of environments (flat surfaces, uneven surfaces, narrow and confined areas, water, and air) [29, 39, 65], in different ways such as walking (on two or more legs), climbing (stairs, racks, etc), maneuvering, traversing, rolling, crawling, and gliding. Communicating status, progress, alerts, feedback, and other information [9, 12, 17, 39] through various means, like speech, display, sounds, gestures, expressions, visual signals, and haptic feedback. And manipulating objects [16, 59, 66] by grasping, lifting, carrying, releasing, pushing, pulling, rotating, and aligning them.
Depending on the work context, this last layer of capabilities can be tuned and tailored to the specific tasks explored, taking into account the capabilities the workers must have to perform it.

Lastly, the realization of these capabilities in a real-world context is influenced by and dependent upon contextual conditions that either enable or constrain their actuation.

# 4. Framework Exploration and Validation

To test and improve the framework, we carried out three related activities. First, we applied it to a range of commercially available, industry-oriented robots to show that (a) the framework captures key aspects of robotic systems and (b) robotics experts tend to produce consistent descriptions when using it (Section 4.1.2). Second, we introduced the framework to students without a background in robotics to evaluate how easily it could be learned. This exercise showed that it not only required minimal instruction but also helped students develop a richer vocabulary and clearer understanding when designing potential robots (Section 4.2.2). As each of these activities relies on a different methodology, we present the approach and outcomes for each separately, aligning the results with the design criteria outlined in Section 3.2.

## 4.1 Representing existing robots using the framework

### 4.1.1 Method

In this section, we elaborate on the results of the activity carried out by robotics experts, who were tasked to describe four robots that can be applied to industrial situations and are representative of our context of research by using the framework: i) Digit, by Agility Robots; ii) SkyPod, by Exotec; iii) ANYmal, by ANYbotics; and iv) Stretch, by Boston Dynamics. In this way, we try to demonstrate that it can provide sufficient knowledge about the current state-of-the-art of robotics. The five robots were chosen as broadly representative of a few different robotic archetypes. Three researchers coded each robot to ensure that there was a good level of consistency in the application of the framework. Areas of ambiguity or disagreement were noted, and discussed.

### 4.1.2 Results

Across the five robots, the three experts converged on a shared, high-level picture of what each platform can do in practice. Mobile and logistics-oriented systems were consistently read as capable of handling goals and executing well-bounded tasks, while manipulation-centric platforms were described as reliably moving and grasping objects according to the operations they have to carry out. For instance, the Stretch robot was uniformly characterized around routine box handling, and both it and the Digit platform were described as working toward pre-defined goals, with the ability to unfold these into sub-goals during execution. Several robots were also recognized as able to change behaviour reactively, without this being interpreted as higher-level autonomy. Equally important, there was convergence on what is not present in day-to-day deployment: learning was generally judged absent, and reasoning beyond the immediate task was often considered outside scope. Read together, these alignments suggest the framework supports experts in rapidly outlining credible capability profiles for state-of-art robots, in a form that can be carried into task-level design discussions.

Where divergence appeared, it often clustered at the boundaries of capability categories. Experts differed on the threshold between "making decisions" and "changing behaviour" for the Stretch robot, for instance, indicating hesitance to credit decision-making when selection is routine or pre-structured. "Move" capability was read variously as locomotion, navigation, or even movement used for interaction: for Stretch and Digit, heterogenous entries reveal that the framework labels can stretch across different interpretations. "Communicate" exposed a line between human-interpretable feedback and back-end signalling. For one expert, ANYmal's and SkyPod's light signals counted as communication, while for the others they did not, treating signals confined to orchestration systems as insufficient. For the "sense" capability a tension between breadth and target specificity was raised whether particular human-oriented detections are required to mark the capability as present. Finally, for some, "learn" meant acknowledging machine learning in the stack and while others needed to see adaptations that occurred during use.

In practical terms, this exercise indicates the need for short, operational prompts that travel with the framework in order to provide better clarity around the scope of each capability.

Table 1: Representation of four commercial industrially oriented robots through the capabilities framework. This description was independently carried out by three researchers for each robot.

| Robot | Capability layer | Capability | Participant 1 | Participant 2 | Participant 3 |
|---|---|---|---|---|---|
| **Digit (Agility Robots)** <br> A bipedal humanoid robot that can be used for parts transport, tool delivery, and warehouse logistics. | *Analysis* | *Sense* | Its own body pose and force, the environment, including objects. No humans or animals yet. | Yes: encoders, cameras, maybe microphones. | It can sense its environment by using Lidar and RBGD camera, and an IMU and motor encoders. With these sensors, it should be able to get a clear picture of its environment and what is happening, and its internal state. Lidar also makes sure it is more robust, at least for navigation, to light conditions. Depending on motor torque sensing, it would also be able to sense interactions with its environment and allow for safe control over the environment and humans. |
| | | *Reason* | No | No | No |
| | | *Change behaviours* | Unspecified but probably not | Limited: it can avoid collision with a dynamic obstacle. | Yes, but very low level (i.e. avoid obstacle) and not complex behaviors due to a lack of reasoning. |
| | *Planning* | *Handle goals* | Predefined goals | Yes | Yes, it can handle predefined goals, high-level goals, and adaptively generate and execute a series of underlying sub-goals |
| | | *Learn* | As far as I can find, not on the "customer" side. The control algorithms are learned but pre-trained. | No | No, it is collecting data, which is used by engineers to update behaviors, so no self-supervised learning. Most things I find about learning are still in simulation or in collaborations with NVIDIA or universities for research. |
| | | *Make decision* | Only path planning (obstacle avoidance), the rest of the planning is done by a human operator. | No | It can make low level descisions based on sensor inputs, like obstacle avoidance |
| | *Interacting* | *Move* | No movement interaction | Yes | From what I'm seeing, I have the feeling it is only able to move robustly in predictable environments and on flat surfaces. |
| | | *Communicate* | Visual: lights (eyes) | Yes: speakers, light | LED visuals, body language (posture to show intentions) and general diagnostics information through an interface aimed at professional operators |
| | | *Manipulate* | All | Yes | Able to manipulate several payloads up to a certain weight, applying a strategy of (manually) interchangeable end-effectors. |
| **SkyPod (Exotec)** <u>A mobile robot designed to move bins between storage racks and picking stations in</u> | *Analysis* | *Sense* | Unknown, but seems very limited. Based on videos, it looks like the AMRs are just following lines painted on the floor. | Yes, encoders | RBG and Laser sensors for navigation purposes |
| warehouses. | | | | | |
| | | Reason | No | No | Reason about abstracting high-level goals into a sequence of concrete tasks according to them. Not sure if I would consider that true reasoning, or if it is just rule-based behavior. Same for reasoning about an optimal strategy in orchestration with the other robots. |
| | | Change behaviours | No | Limited: it can avoid collision with a dynamic obstacle | Yes it can change its behavior based on internal sensing, but also in orchestration with the other robots to optimize some goal. |
| | Planning | Handle goals | No, all robots seem to be managed from a central fleet management system. | Yes | Handle high-level pre-defined goals |
| | | Learn | No | No | Learn from data and distill that into a pre-trained model updates, no real-time learning |
| | | Make decision | No | No | Yes it can make decisions not only based on sensor output, but also based on the states of others robots in the swarm and some common goal (max efficiency) |
| | Interacting | Move | No interaction | Yes | Move in 2D space and interact with storage racks. Requires flat surface floors |
| | | *Communicate* | Seems very basic, a light to show it's on. | Yes: light | No form of direct communication to humans, but again, status reports through a professional interface. Communication with its fellow robots |
| | | *Manipulate* | Grab totes from specific shelves and move them. | Limited: pushing | Grasping, carrying, and lifting standardized bins |
| **ANYmal (ANYbotics)** <u>A quadruped robot that can perform routine visual, thermal, or acoustic inspections of aircraft engines, landing gear, and undercarriage areas.</u> | *Analysis* | *Sense* | Its own body pose and force, environment including: objects, temperatures, sounds. Unclear if it can detect and move around humans. | Yes: encoders, cameras, maybe microphones | RGB-D camera, thermal camera, ultrasonic sensing, with optionally gas sensing and 3D scanners (for inspection). Also torque-controlled elastic actuators (so also torque sensing) and encoders |
| | | *Reason* | Yes, to a certain extent. It can interpret the pose of valves and human-readable dials | No | Difficult, I believe not, since this is a very open platform where reasoning could be implemented by clients |
| | | *Change behaviours* | Can handle "multiple missions" | Limited: it can avoid collision with a dynamic obstacle | Low level for navigation and preventing from falling on uneven terrains |
| | *Planning* | *Handle goals* | Pre-defined and online updated goals (in the form of better/more optimal trajectories) | Yes | I would say medium-level goals, like go from A-B |
| | | *Learn* | As far as I can find, not on the "customer" side. The control algorithms are learned, but pre-trained. For the first deployment, "The robot is taught the layout of the workplace", but this is between quote marks on the site, and seems to refer to SLAM instead of (reinforcement) learning. | No | No real-time learning, only pre-trained models from data. |
| | | *Make decision* | Yes, planning routes and orders for missions | No | Low-level decision-making that is necessary, for example, for navigation |
| | *Interacting* | *Move* | Unclear but probably not | Yes | Yes |
| | | *Communicate* | Lights and sounds | Yes: speakers, light | Only through some operator control interface |
| | | *Manipulate* | No | Limited: pushing | No manipulation of the environment |
| **Stretch (Boston Dynamics)** <u>A mobile warehouse robot designed to automate case handling tasks.</u> | *Analysis* | *Sense* | Its own body pose and force, environment, including objects. No humans or animals | Yes: encoders, cameras | |
| | | *Reason* | No | No | Reason about the high-level task, given an understanding of the scene to think about the task sequencing, motion, and grasp planning |

| | | Change behaviours | No | Limited: it can avoid collision with a dynamic obstacle | Low level navigation |
| | Planning | Handle goals | As far as I can find only "Load" or "Unload" | Yes | Pre-defined high level |
| | | Learn | No | No | Pre-trained models, no real-time adjustments |
| | | Make decision | The decision on which box to pick up next | No | Low-level |
| | Interacting | Move | No interaction | Yes | 2D space requiring a flat surface and semi-structured environments. |
| | | Communicate | No | Yes: light | Handheld pendant (controller), alarm sounds and communication with its fellow robots |
| | | Manipulate | Pickup and move boxes | Yes | Grasp, lift, and carry a wide variety of case types |

## 4.2 Conceptual design with the framework

To explore how the framework can support the conceptualization of robotic agents, we conducted a light design session (45 minutes) in an interdisciplinary course exploring HRI at [anonymous university]. The goals of the course are to learn about robotics and HRI, and to develop prototypes of robotic platforms, adhering to a client brief and working in subgroups of four to five students.

At this point in the course, the 35 students had already undertaken classes around robotics from different disciplinary perspectives, and were in the conceptualization phase of their prototypes. Before the session, we briefly introduced the robotic capabilities to the students in order for them to familiarize with it.

### 4.2.1 Design Task

The design task we asked the students to carry out consisted of a multistep exercise, in which they had to situate their conceptual robots in the real-world workplace, define their initial characteristics, sketch the robots, and reflect on their capabilities. The exercise was supported with a very simple tool that translates the knowledge instilled in the framework and acts as a guide to better detail their concepts in terms of robotic capabilities.

The tool consists of several sheets (Fig. 3) through which:
- Develop an initial story in which their robots have been introduced to the workplace and they are assisting workers in different tasks.
- Specify the robot's identikits by defining their aspect, degree of movement, autonomy, and collaboration, working environment, and aim, through a point system (from 0 to 10)
- Sketch their concepts (not mandatory).
- Define the robot's capabilities using our framework.
- Complete the initial story by mentioning what the robot and the workers do, how they collaborate, and what kind of relationship they establish with each other.

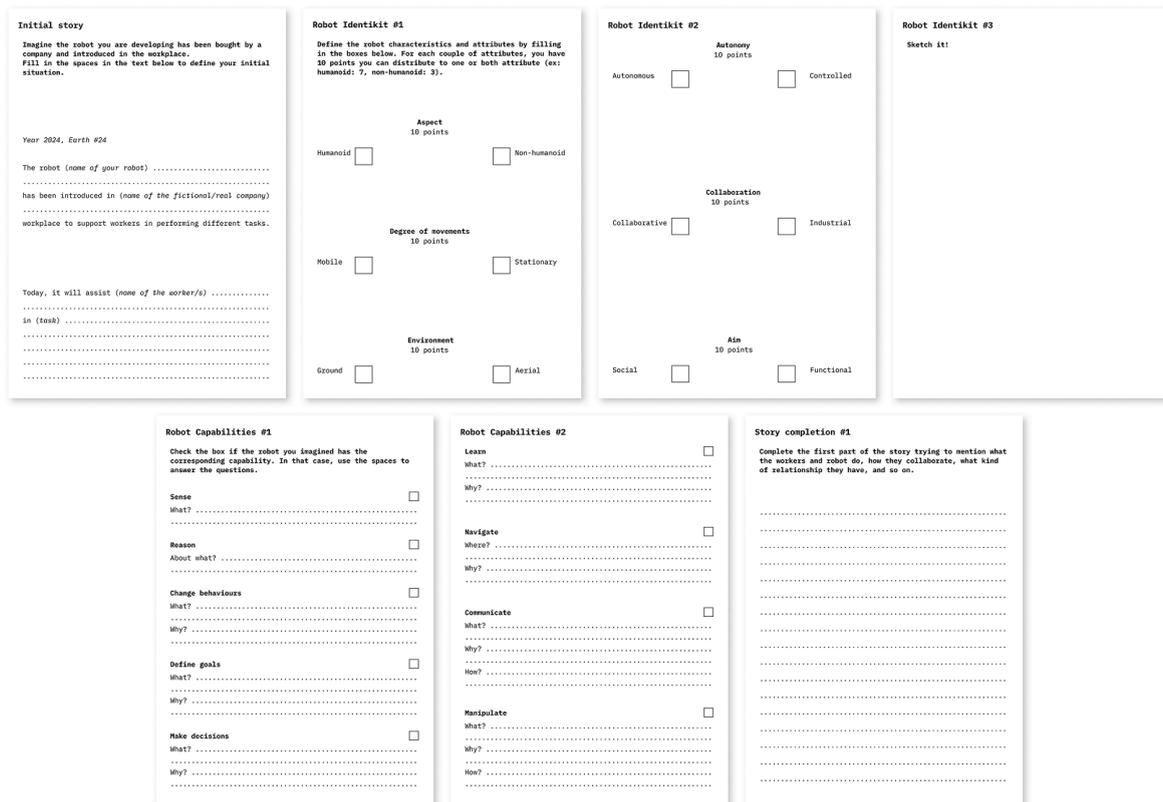

Fig. 3 The exercise sheets provided to the students. From left to right, we find: i) the sheet to define the initial story; ii) two sheets to define the robot characteristics; iii) the sheet where to sketch the robot's appearance; iv) two sheets to define the capabilities the robot should have; and v) the sheet to complete the initial story.

As this work was carried out in an educational capacity, we are not presenting it as a formal validation of the framework, as the data would not support a systematic analysis. Rather, we use it to *illustrate* the way that the framework can be used, and demonstrate that it is meaningfully comprehensible in a limited period of time to students who are not experts in robotics. We do this using i) an illustrated example of how the framework was applied by one of the student groups, supported with direct quotes from their work and ii) a reflexive analysis based on field notes that describes the progress of the cohort as a whole.

### 4.2.2 Illustrated outcome

To guide the reader through this activity, we present the work developed by one of the groups as an illustrative example. We follow the same textual structure that was provided to the students (with the group's input shown in *italics*).

"Year 2024,
The *Robottier* has been introduced in *a pediatric hospital* to support workers in performing different tasks. Today, it will *assist nurses and will help to involve parents in preparation of doses of baby feed, that is collecting breast milk, taking prescriptions, mixing additional fortifiers into the milk, dosing and heating it to the correct temperature to be given to the patients.*"

Concerning the robot's appearance (humanoid vs non-humanoid), the group conceptualized it as non-humanoid, assigning the maximum score of 10 points. In terms of mobility, the robot was described as predominantly stationary (8 points out of 10). As for the operational environment, the group clearly indicated it is a ground robot (10 points).

The robot was also expected to demonstrate a range of autonomous behaviors, scoring 7 out of 10. It was characterized as highly collaborative, receiving 9 out of 10 points. Its intended goals were both social (4 points) and functional (6 points).

In terms of capabilities, the group identified the following: the robot should be able to sense *objects, detect users' actions, and monitor power availability*. It should adapt its behavior *according to the timing of specific tasks and manage access to compartments based on the actor involved, allowing for personalized interaction with different patients and nurses*. The robot should make decisions regarding *task planning and navigation paths to retrieve necessary components and materials, taking into account feeding schedules, prescriptions, and available resources*. It should also communicate *its operational status, errors, and action prompts to users in order to reassure both parents and nurses, ensure safe functioning, and maintain workflow continuity*. Lastly, the robot should be capable of manipulating *breast milk and packaging it into individual doses, thus reducing the number of repetitive tasks performed by nurses*.

The group completed the initial narrative as follows: "*all prescriptions for the day are retrieved and verified. The robot measures and dispenses the correct amount of fortifiers. After the final dose from the previous day has been delivered, the robot prompts the nurse to refill the syringe storage, clean any spillage, and sanitize contact points. It then instructs the nurse to add the dispensed fortifier into a measured quantity of milk. The nurse places the filled container into a designated holder, enabling the robot to access it. From this point onward, the robot autonomously prepares each dose and notifies the nurses only when the dose is heated and ready to be administered. At any time during the day, parents can deposit breast milk into a refrigerated compartment within the robot.*"

Despite the limited time available (15 minutes for the explanation of the framework and 45 minutes to complete the exercise), each group showed a strong understanding of the framework and its terminology, requiring no further clarification during the hands-on session.

They used terms from five capabilities' groups out of the nine proposed through the framework, covering a broad range of possibilities. They further connected terms from the framework to the activity of humans around robots, such as assisting nurses in order to reduce repetitive tasks and maintain workflow continuity, allowing them to have a more personal and personalized interaction with the patients.

In Figure 4 we show the sheets filled by the group.

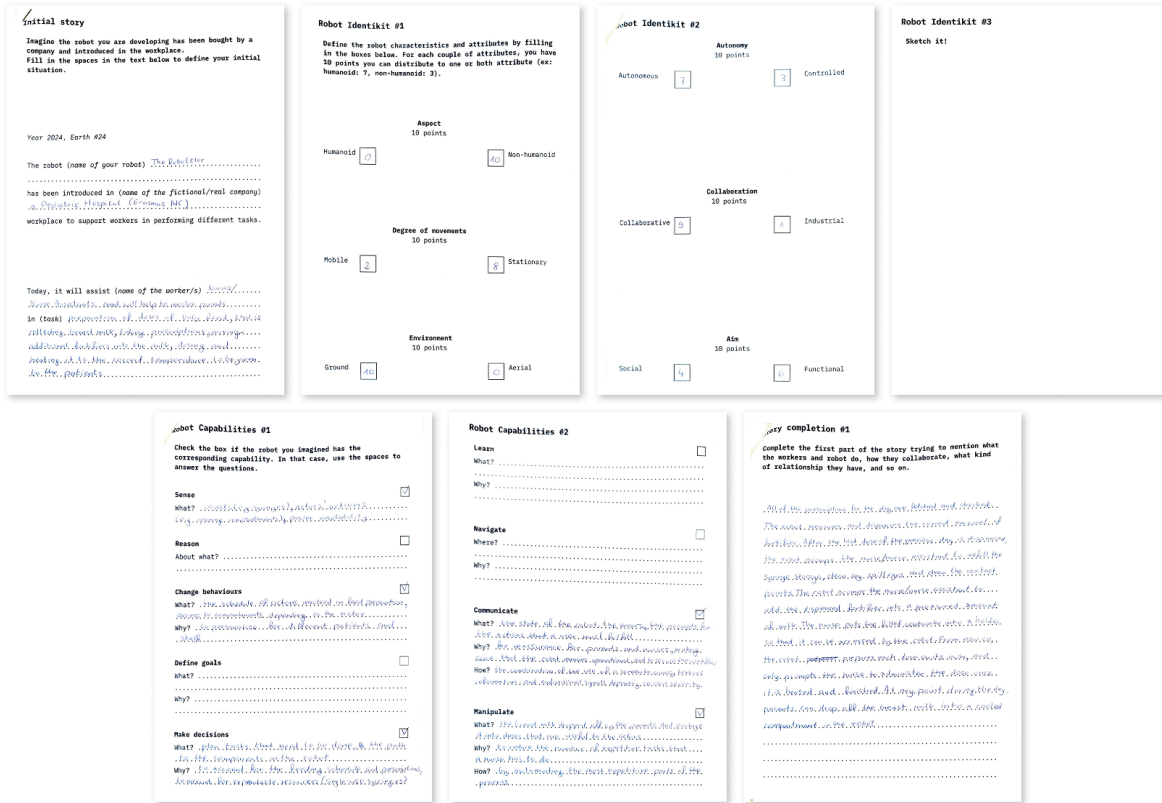

Fig. 4 The sheets filled by the group.

### 4.2.3 Reflexive analysis of educational use

Looking across the student groups at a higher level, and including reflections from the educator, we can note that all the groups successfully completed the task within the allotted time, demonstrating both the accessibility and the clarity of the framework. The graphic materials provided to the students (which can be seen as a visual and practical translation of the framework) proved sufficient for them to apply the framework to their design brief without requiring additional explanation or intervention.

Interestingly, all the groups were able to reframe their concept from a different perspective, shifting their focus from hardware, software, and technicalities toward capabilities, interaction with human roles, and how meaningful human-robot collaboration could be envisioned. This shift indicates that the framework not only guided their design work but also supported a change in mindset, helping them approach robots less as technical artefacts and more as agents situated within human contexts.

We can also observe that the students were able to describe their concepts using key terms drawn directly from the framework's terminology, showing that they had internalized its categories and concepts. In turn, this demonstrates how the framework was assimilated into their ways of thinking and became part of their design language. Additionally, by looking at the groups' conversations during the activity, it is evident that the framework provided a useful level of specificity for their ongoing design process and communication, allowing them to structure discussions more clearly and reduce ambiguity.

Based on this, we conclude that the students quickly assimilated the framework and were able to engage with it productively. The framework offered them a more nuanced vocabulary for discussing robotic concepts, supported their reflection on task allocation between workers and robots, clarified the features required of the robots, and helped them better define and contextualize the operational scenarios they were envisioning.

# 5. Discussion

In this paper we provide a framework for articulating robotic capabilities in a way that makes them accessible to non-experts and allows for creative design work. We discuss in Section 5.1 first how this framework relates to the criteria set out in Section 3.1 and its current limitations (Section 5.1.1). We then discuss the function of this kind of framworks as a boundary object (Section 5.2) and intermediate-level knowledge (Section 5.3), and the wider epistemic import of the practices (Section 5.4).

## 5.1 Framework validation

The robotic capabilities framework was developed according to a set of criteria aimed at ensuring both its practical relevance and its accessibility to non-expert stakeholders (see Section 3.1). In this section, we evaluate the framework against those criteria, drawing on our empirical activities and comparative analysis. For each criterion, we discuss what evidence we have for satisfaction, and nuances of how the work has played out.

*C1*: *The framework captures the range of functions that current and near-future robots may offer, while presenting them in a simplified, accessible format that is understandable even to non-experts, avoiding technical specification.*

This is the criterion that emphasizes *comprehensiveness* and *accessibility*. We tried to satisfy its first part (*the framework captures the range of functions that current and near-future robots may offer*) already in the early stages of the framework development, through the 1-1 discussions, the focus group, and the comparison with the existing literature, verifying that it provides sufficient coverage. Furthermore, the activity carried out by the roboticists shows the framework is able to capture key aspects of current robotic systems in a way that experts can recognize and use to articulate task-level implications. It further shows that experts tend to be consistent where categories are clear and that apparent inconsistencies are localized at the boundaries between capabilities.

The second part of this criterion (*while presenting them in a simplified, accessible format that is understandable even to non-experts, avoiding technical specification*) was partially validated through the design session with the students, most of whom had no technical training in robotics. This session demonstrated that they could confidently understand and apply the framework in a short period of time, affirming its accessibility. This positions the framework as a practical tool for inclusive design conversations without sacrificing descriptive power.

*C2*: *It serves as a boundary object, supporting high-level, transdisciplinary discussions about the integration of robots in work contexts*, *and facilitating communication and collaboration between different stakeholders.*

This is instead the criterion that puts emphasis on *mediation through (shared) language*.

We saw its partial validation during the design session held with the multidisciplinary group of students who were in dialogue with the workers who might eventually adopt the proposed robotic platforms. Throughout the session, students articulated their concepts using the framework's vocabulary, indicating that it had been internalized in their reasoning.

Furthermore, the framework proved sufficiently precise for intra-team communication: it helped them debate and refine ideas, structure reflection, identify required features and capabilities, and situate operations within the scenarios they envisioned. Their engagement suggests the framework can scaffold early ideation and ongoing refinement, shifting attention from robots as technical artefacts to robots as situated agents within human contexts.

**C3**: *It emphasizes what robots can do, highlighting distinctive features that help compare and differentiate robots in meaningful ways.*

Finally, this criterion emphasizes the importance of adopting a *task-oriented perspective*, one that frames robots in terms of their capabilities and the actions or activities they can undertake. This orientation is built into both the framework's structure and terminology, directing attention toward meaningful affordances and practical implications rather than technical specifics.

In doing so, during the design session, the framework not only shaped the students' design process but also encouraged a shift in perspective, enabling them to better recognize which aspects of workers' activities could be meaningfully supported by robots.

This framework capacity became further clear during the classification of robots done by the experts, where the framework revealed differences in features not immediately apparent through traditional descriptions.

### 5.1.1 Limits

Although we see value and relevance in our contribution, we must recognize several limitations that affect our work. We begin this section by illustrating more general and processual flaws to then turn to more technical limits of the framework itself.

By abstracting capabilities into a boundary object, we inevitably omit technical specifics that materially shape capability in practice (e.g., payload limits), which may influence downstream feasibility judgments. Furthermore, our validation was not genuinely transdisciplinary: as a result, we can't assert that the second criterion was fully validated. Again connected to transdisciplinarity, we lack input from key areas of knowledge concerning workers and work contexts (e.g., work psychology, organizational studies), and we have not yet involved workers in contributing to the definition of (desirable) robotic capabilities themselves. Lastly, we are not able to demonstrate that the concepts produced by the students with the framework are "better" in any comparative or outcome-based sense.

Moving to more inherent limits of the framework, we must note that while the framework provides a structured overview of key robotic capabilities, it does not differentiate between the level of maturity of a capability, nor does it address how capabilities may be combined, scaled, or prioritized in relation to one another. The framework also treats capabilities as relatively stable, leaving aside questions of how they might evolve over time or adapt across different tasks and contexts. Although contextual conditions are recognized as enablers or constraints, the framework does not specify how such conditions influence the extent or quality of a capability's realization.

Finally, the analysis of expert usage points to small but consequential adjustments needed to improve the vocabulary, so that task-oriented design conversations can proceed on firmer common ground. Together, these limits suggest opportunities for extending the framework to capture the variability, interdependence, and developmental trajectories of capabilities.

## 5.2 The robotic capabilities framework as a boundary object

The robotic capabilities framework operates as a boundary object in the sense articulated by Star and Griesemer [61]. Throughout the development and application activities described in Sections 3 and 4, the framework served precisely this mediating function. It enabled actors situated in different epistemic cultures to engage with a shared representation of what robots can do, while allowing each group to interpret and elaborate it according to its own concerns. This boundary-object function emerged in three concrete ways.

First, the framework provided a shared vocabulary for discussing robotic capabilities without requiring agreement on deeper disciplinary assumptions. Robotics experts can engage with the framework as a high-level structuring of functional competencies; designers may use it as a scaffold for reasoning about interactions and relationships; and students might benefit from it to articulate conceptual roles for novel robots. Despite these different orientations, the framework maintains a stable referential structure across groups, mirroring what Akkerman and Bakker describe as a boundary object's ability to support coordination without consensus [1].

Second, the framework's emphasis on task-oriented capability descriptions allows to translate between technical and experiential knowledge. Because capabilities are framed in relation to tasks (Section 3.3), stakeholders with little robotics expertise can meaningfully contribute to discussions about what roles robots might assume, how tasks can be reconfigured, and where human expertise is irreplaceable. This resonates with prior HCI and design literature on boundary objects as mediators of collaborative sensemaking that bridge heterogeneous knowledge systems [43, 49].

Third, the framework supports productive ambiguity, an attribute often central to boundary objects [38, 61]. During expert coding of existing robots (Section 4.1), mild divergences appeared around category boundaries, such as whether light signals constitute "communication", illustrating that the framework leaves room for interpretation while still enabling alignment. These divergences don't hinder collaboration; rather, they help surface tacit assumptions, prompting clarification and negotiation across disciplinary boundaries. As Klein argues, such boundary interactions are crucial for transdisciplinary work, enabling integration without homogenization [35].

Our framework allows diverse stakeholders to work with the same conceptual tool while preserving the situatedness of their respective knowledge practices, which is an essential requirement for responsible and participatory innovation in HRI [19, 56, 75].

## 5.3 The robotic capabilities framework as intermediate-level knowledge

Beyond its role as a boundary object, the robotic capabilities framework also functions as intermediate-level knowledge in the sense established by Höök and Löwgren [25, 41]. The framework fulfills this role by providing a design-relevant abstraction of robotic capabilities that is neither overly general nor tied to a specific platform or work domain. The framework exhibits several characteristics typical of intermediate-level knowledge.

First, it offers a structured, transferable articulation of robotic capabilities that can be taken up in new design processes. It does not prescribe design solutions, nor does it aspire to a universal theory of robotics; instead, it provides a vocabulary and conceptual organization that designers and engineers can adapt, extend, or reconfigure. This aligns with the role of strong concepts as generative design elements that can seed new design explorations across projects and settings [25].

Second, the framework is grounded in empirical and material practices. Its development was informed by expert discussions, iterative synthesis (Section 3.1), robot profiling (Section 4.1), and design exercises with students (Section 4.2). Löwgren emphasizes that intermediate-level knowledge must be derived from and validated through design practice, rather than remaining purely theoretical [41]. The framework satisfies this criterion by demonstrating usability and conceptual traction across multiple empirical activities.

Third, it demonstrates a useful degree of abstraction. The capability categories are high-level enough to accommodate a wide variety of robots, yet specific enough to support concrete design reasoning around task allocation, interaction possibilities, and collaboration patterns. This makes the framework actionable for early-stage concept development, participatory workshops, or preliminary feasibility discussions [20, 25].

Fourth, the framework is positioned to support reflection and knowledge accumulation in HRI, an increasingly recognized need within design-oriented HRI research [42]. By offering a structured representation of capabilities, it enables researchers to compare concepts across projects, articulate recurring patterns in human–robot collaboration, and reason more systematically about how capability configurations shape relational dynamics. In this sense, the framework contributes to building a shared, transdisciplinary knowledge base in HRI, supporting the maturation of the field.

Finally, because the framework is intentionally adaptable, it can evolve through future use, potentially becoming part of a larger repertoire of design tools for HRI.

In sum, the robotic capabilities framework provides a generative structure that supports the design of meaningful worker–robot relations while also enabling knowledge transfer across disciplinary and practical boundaries. As such, it contributes not only to specific workplace applications but also to the broader epistemic development of design-oriented HRI.

## 5.4 Epistemic contributions and reflections

Given our broader aim, we will further elaborate on the epistemic contributions that each area of knowledge has provided or can potentially offer to this work. From a knowledge integration perspective, robotic engineering contributes by offering in-depth technological knowledge and a technical understanding of robots and their capabilities. This expertise helps mitigate unrealistic expectations about technological integration [60] by facilitating the transfer of accurate and actionable knowledge to workers. Roboticists typically adopt a functional lens, emphasizing performance outcomes that can improve working conditions and enable meaningful human-robot collaboration [21]. However, they often find it difficult to address the relational dimensions that intersect with these functional concerns.

Design researchers in our team have played a complementary role by translating disciplinary knowledge about robots and their potential into visual and material forms that resulted in the framework. Conversely, they tend to focus on the experiential and relational aspects of human-robot interaction, drawing attention to broader socio-technical issues. Yet, they may lack the deep technological understanding needed to engage with the complexity of robotic systems. These disciplinary differences became especially evident in our discussions around the term "*capability*", where conceptual frictions arose. Although such tensions highlight epistemological challenges, they also underscore the necessity of developing a shared and accessible vocabulary, such as our robotic capabilities framework.

# 6. Conclusion

In this paper, we elaborate on a robotic capabilities framework as a shared vocabulary for describing what current and near-future robots can do. The framework presents capabilities in a format accessible to non-experts, supports conceptual and transdisciplinary discussions about integrating robots into work contexts, and functions as both a boundary object and an intermediate-level form of knowledge. It emphasizes what robots could be capable of, highlighting distinctive features and enabling comparison across platforms.

We propose this framework to address a gap in HRI: the lack of tools that effectively support multistakeholder conversation and collaboration around robot integration in the workplace.

We conducted an initial, partial validation through two activities: (i) an exercise in which domain experts profiled existing robots using the framework's terminology, and (ii) a short workshop in which multidisciplinary students applied the framework to conceptualize robotic platforms for specific work contexts.

Findings indicate that the framework is readily understood and usable even without prior HRI or robotics expertise, positioning it as a practical aid for HRI discourse. It also scaffolds early ideation and iterative refinement, encouraging teams to treat robots as situated agents within human contexts and helping them identify aspects of work that can be meaningfully supported by robotic capabilities.

These results are promising and point toward several opportunities for future work. Expanding the framework with broader disciplinary and practice-based input, as well as refining how it captures variability, interdependencies, and the evolution of capabilities over time, will strengthen its relevance. By doing so, our framework can mature into a tool that not only supports rigorous analysis but also empowers diverse stakeholders to collaboratively imagine and shape more meaningful, equitable, and effective integrations of robots in the workplace.

# Acknowledgements

This work was supported by the BrightSky project, funded by the R&D Mobiliteitsfonds from the Netherlands Enterprise Agency (RVO) and commissioned by the Ministry of Economic Affairs and Climate Policy.

We want to thank the members of our research team who engaged to different extent with this research, making it possible: J. Micah Prendergast, Luka Peternel, Auke Nachenius, and Stephan Balvert.